# Comprehensive Implementation of TextCNN for Enhanced Collaboration between Natural Language Processing and System Recommendation

Xiaonan Xu[1,*] Zheng Xu[2], Zhipeng Ling[3], Zhengyu Jin[4], ShuQian Du[5]

[1*]Independent Researcher, Northern Arizona University, Flagstaff, USA
[2]Computer Engineering, Stevens Institute of Technology, hoboken, NJ, USA
[3]Computer Science, University of Sydney, Sydney, Australia
[4]Informatics, Univeristy of California, Irvine, Irvine, CA, USA
[5]Information Studies, Trine University, Phoenix, Arizona, AZ, USA

*Corresponding author: xiaonanxu5@gmail.com

**ABSTRACT** Natural Language Processing (NLP) is an important branch of artificial intelligence that studies how to enable computers to understand, process, and generate human language. Text classification is a fundamental task in NLP, which aims to classify text into different predefined categories. Text classification is the most basic and classic task in natural language processing, and most of the tasks in natural language processing can be regarded as classification tasks. In recent years, deep learning has achieved great success in many research fields, and today, it has also become a standard technology in the field of NLP, which is widely integrated into text classification tasks. Unlike numbers and images, text processing emphasizes fine-grained processing ability. Traditional text classification methods generally require preprocessing the input model's text data. Additionally, they also need to obtain good sample features through manual annotation and then use classical machine learning algorithms for classification. Therefore, this paper analyzes the application status of deep learning in the three core tasks of NLP (including text representation, word order modeling, and knowledge representation). This content explores the improvement and synergy achieved through natural language processing in the context of text classification, while also taking into account the challenges posed by adversarial techniques in text generation, text classification, and semantic parsing. An empirical study on text classification tasks demonstrates the effectiveness of interactive integration training, particularly in conjunction with TextCNN, highlighting the significance of these advancements in text classification augmentation and enhancement.

**Keyword list** Natural Language Processing (NLP);Text classification-CNN;Deep learning;Semantic parsing

Introduction

Natural language processing (NLP), an important branch of artificial intelligence, has developed rapidly in recent years, driven by deep learning. Deep neural networks, such as recurrent neural networks (RNNS) and convolutional neural networks (CNNS), have demonstrated excellent performance in semantic understanding, speech recognition and machine translation, but they also face security risks such as anti-sample and poison attacks. In many cases, the diversity, complexity and distribution of information in text form makes traditional manual analysis methods ineffective, forcing people to look for new solutions. Intelligence texts cover a wide range of media and sources, including news reports, spy reports, battlefield reports, government documents, social media posts and more. These texts can contain information from all parts of the world and at all times, making them highly timely and valuable. However, this diversity also presents significant challenges, as texts from different sources may use different formats, languages and phrases, leading to fragmentation of information and increased difficulty in analysis. In addition, intelligence texts themselves are often highly specialised, containing a large number of technical terms, acronyms and domain-specific languages, which places higher professional demands on analysts and therefore requires a more intelligent and efficient approach to this problem.

Traditional text classification methods generally need to pre-process the text data of the input model, in addition to obtaining good sample features through manual annotation, and then use classical machine learning algorithms to classify them. Similar methods include Naive Bayes (NB), K-Nearest Neighbour (KNN), Support Vector Machine (SVM), etc. The level of feature extraction has an even greater impact on text classification than on image classification, and feature engineering in text classification is often time consuming and computationally expensive. After 2010, the method of text classification gradually shifted to deep learning model. Deep learning applied to text classification maps feature engineering directly to the output by learning a series of nonlinear transformation patterns, thus integrating feature engineering into the process of model fitting, and has achieved great success once applied. Therefore, this paper delves into the current state of applying deep learning to the fundamental tasks of Natural Language Processing (NLP), which encompass text representation, word order modeling, and knowledge representation. It further explores strategies for enhancing natural language processing technology, specifically through the utilization of text classification algorithms, in response to the challenges posed by attack methodologies in text generation, text classification, and semantic parsing. The effectiveness of an integrated interactive training approach is corroborated through empirical studies conducted on text classification tasks. This comprehensive analysis integrates insights from "Classification Augmentation and Synergy through TextCNN and Natural Language Processing," underscoring the significance of these advancements in the NLP field.

## 2. Application of Deep Learning in Natural Language Processing

With the continuous development of computer and artificial intelligence, Natural Language Processing (NLP) technology, as one of the research directions, has been applied in more and more fields, including text processing, speech recognition and translation, semantic understanding and knowledge-based text mining.

### 2.1 Natural language processing concepts

In September 2019, the Allen Institute for Artificial Intelligence (AI2) released a computer program called Aristo that correctly answered more than 90 per cent of the questions on an eighth-grade science test. Passing a high school exam may sound trivial, but it's complicated for a computer. Aristo used natural language processing (NLP) to find answers from billions of documents. NLP is a branch of computer science and artificial intelligence that enables computer systems to extract meaning from unstructured text. While computers are still a long way from being able to understand and use human language, NLP is already key to many applications we use every day, including digital assistants, web search, email and machine translation.

### 2.2 Natural language - text preprocessing

Text preprocessing is a very important step in NLP, its goal is to convert raw text into a form that a computer can process. Common preprocessing operations include word segmentation, stop and stop words, part-of-speech tagging and so on. Word segmentation refers to the division of sentences into single words, the elimination of words that have a high frequency in the text but have no meaning, such as "the", "a", etc., part of speech tagging refers to the assignment of each word to a part of speech, such as nouns, verbs, adjectives, etc. These actions can help the computer better process text data.

Pre-trained language models can effectively learn global semantic representation and significantly improve the effectiveness of NLP tasks, including text classification. It typically uses an unsupervised approach to automatically mine semantic knowledge and then builds pre-trained targets that allow the machine to learn to understand the semantics. As shown in Figure 1, it is assumed that the differences between Embeddingfrom Language Model (ELMo), OpenAI GPT, and BERT are given. ELMo is a deeply contextualized word representation model that can be easily integrated into models.

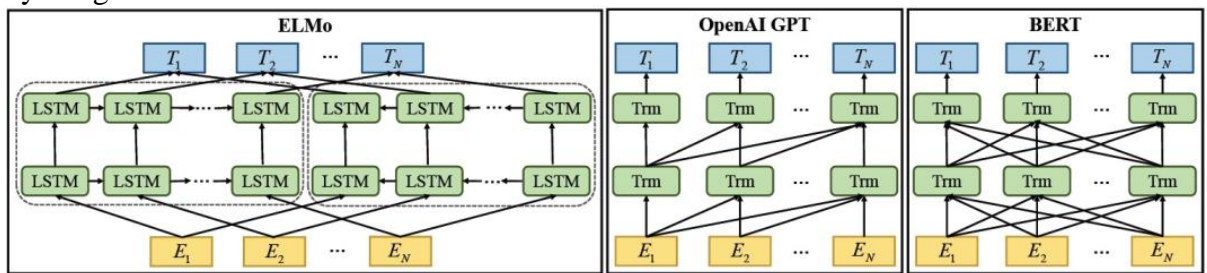

Figure 1: Classification of natural language processing models

Among them, preprocessed text can model the complex features of words and learn different representations for various linguistic environments. It learns the embeddings of each word based on the context of the bidirectional LSTM. GPT employs supervised fine-tuning and unsupervised pre-training to learn general representations that transfer to many NLP tasks with limited adaptations. In addition, the domain of the target dataset does not need to be similar to the domain of the unlabeled dataset. The training process of GPT algorithm usually includes two stages. First, the initial parameters of the neural network model are learned by modeling objectives on an unlabeled data set.

### 2.3 Text classification-CNN

Text classification is an important problem in NLP, where the goal is to classify a given text into predefined categories. Common algorithms include naive Bayes, support vector machines, deep learning, and so on. Naive Bayes is a classification algorithm based on probability statistics, which assumes that features are independent of each other; support vector machine is a classification algorithm based on maximum boundary separation, which can map nonlinear separable data to high-dimensional space for classification; Deep learning is a classification algorithm based on neural networks, which can automatically extract features and classify them.

TextCNN is an algorithm that uses Convolutional Neural Networks to classify text. It was proposed by Yoon Kim in 2014 in the article Convolutional Neural Networks for Sentence Classification. The detailed schematic diagram is as follows.

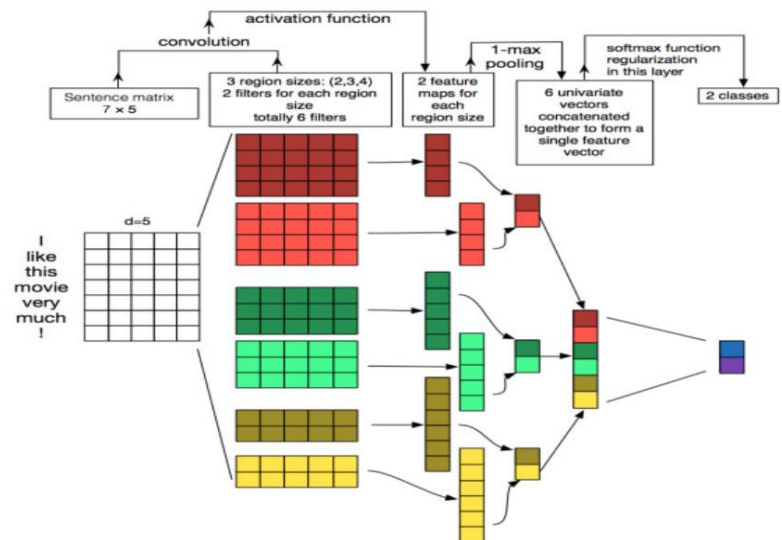

Figure 2: Schematic diagram of text-CNN Text classification model

The input dimensions are: [batch_size, seq_len, emb_dim]. In the example above, batch_size=1, seq_len=7, emb_dim=5;

In the convolution, it is actually a single channel, the width of the convolution kernel is the same as the dimension of the word vector, which is emb_dim, and the height of the convolution kernel is specified by itself;

Because the width is emb_dim, the convolution can only move up and down, not left and right. In the figure above, there are two convolution kernels of high dimension 4, two convolution kernels of high dimension 3, and two convolution kernels of height 2.

The feature map after each convolution is downsampled with maximum pooling.

In the example above, it is equivalent to having 3 convolution layers, each of which has 2 convolution nuclei, then the convolution layers with sizes 2, 3 and 4 have the following dimensions: [batch_size, 2, seq_len-1, 1], [batch_size, 2, seq_len-2, 1], [batch_size, 2, seq_len-3, 1];

The dimensionality of each layer after convolution pooling is: [batch_size, 2], where 2 is the number of convolution nuclei;

The results after maximum pooling are splicing in the following dimensions: [batch_size, len(filter_sizes) * num_filters], and fed into the full connection for classification.

filter_sizes are the heights of the convolution cores and num_filters are the number of convolution cores.

### 2.3 Bi-LSTM text classification application

LSTM: Long-term short-term memory network (LSTM) can remember the information before and after long sentences, solving the problem of RNN (when the time interval is large, the network will forget the previous information, resulting in the problem of gradient disappearance, which will form a long-term dependency problem) and avoiding long-term dependency problems.

Therefore, the combination of Bi-LSTM and TextCNN has several advantages for emotional text classification. First, Bi-LSTM (Bidirectional Long Short-Term Memory) can effectively capture contextual information in the text because it considers both forward and backward sequences of the text to better understand the relationship and context between words in the text. The ability of the model to capture the long-range dependency relationship is improved. Compared to traditional recurrent neural networks (RNN), Bi-LSTM mitigates the gradient disappearance problem and makes the model easier to train.

$$\begin{cases} \overrightarrow{h_t} = \text{LSTM}(x_t, \overrightarrow{h_{t-1}}) \\ \overleftarrow{h_t} = \text{LSTM}(x_t, \overleftarrow{h_{t-1}}) \\ h_t = \omega_t \overrightarrow{h_t} + \vartheta_t \overleftarrow{h_t} + b_t \end{cases} \quad (1)$$

Where :ω and is the weight coefficient; b is a biased value. Compared with a single LSTM model, the Bi-LSTM model not only takes into account the forward correlation information between the time series data, but also takes into account the reverse correlation information between the time series data, so it shows superior performance in the classification of sequence data.

In this paper, TextCNN is used to embed words, which are then fed into the Bi-LSTM network for fine tuning to complete parameter optimisation for text classification tasks. Bi-LSTM reads the text sequence X from both directions and computes the hidden state of each word, then concatenates the hidden states in both directions to obtain the final hidden representation of the ith word:

$$h_i = [\overrightarrow{h_i}; \overleftarrow{h_i}] \quad (2)$$

Therefore, the process by which the Bi-LSTM network obtains the final feature vector output by learning BERT input data can be expressed as:

$$F = f_{\text{Bi-LSTM}}(f_{\text{BERT}}(s; \theta_{\text{BERT}}); \theta_{\text{Bi-LSTM}}) \in \mathbf{R}^D \quad (3)$$

Where: indicates the mapping relationship; s represents the text sentences, TextCNN and parameters of the Bi-LSTM network :D represents the dimensions of the hidden layer in the Bi-LSTM network.

Therefore, TextCNN (Convolutional Neural Network for Text) has the ability to perform convolutional operations on different window sizes, allowing the model to capture features at different scales in the text, from local to global. This helps to extract key information and features from the text, making the model more robust and able to adapt to text of different lengths and structures.

In summary, by combining Bi-LSTM and TextCNN, we can take full advantage of their respective strengths and effectively tackle emotional text classification tasks. Bi-LSTM is responsible for capturing the contextual information of the text, and TextCNN is responsible for extracting features at different scales. The two complement each other, improving the performance and generalisation ability of the model to achieve better results in emotional text classification. This combined approach has already achieved remarkable success in several natural language processing tasks.

### 2.4 TextCNN Procedure

According to the TextCNN principle architecture in Figure 2, it can be seen,

(1) The first layer is input layer.

$$L_{out} = \left\lfloor \frac{L_{in} + 2 \times \text{padding} - \text{dilation} \times (\text{kernel\_size} - 1) - 1}{\text{stride}} + 1 \right\rfloor \quad (4)$$

In the formula, it is assumed that the dimension of the input data is 1, the length is 8, the number of channels is 1, and the dimension and number of filter are 1 and the length is 5. The input layer is an n× k matrix where n is the number of words in a sentence and k is the dimension of the word vector corresponding to each word. That is, each row of the input layer is the K-dimensional word vector corresponding to a word. In addition, padding is applied to the original sentence to make the length of the vectors consistent.

(2) The second layer is the convolution layer, and the third layer is the pooling layer. First, we should note the differences between convolution operations in computer vision (CV) and NLP. In CV, the convolution kernel is usually square, such as the 3×3 convolution kernel, and then the convolution kernel moves along the height and width of the entire image to carry out the convolution operation. Different from CV, the "image" of the input layer in NLP is a word matrix composed of word vectors, and the width of the convolution kernel is the same as the width of the word matrix, which is the size of the word vector, and the convolution kernel only moves in the height direction.

$$x_i$$
$$x_i \in \mathbb{R}^k \quad (5)$$

$$x_{i:j}$$
$$x_{i:j} = x_i \oplus x_{i+1} \oplus \ldots \oplus x_j \quad (6)$$

Therefore, the position that the convolution kernel slides through each time is a complete word, and the one-minute "vector" of several words is not convolved.

(3) Finally, there is the pooling layer. The network shown in the figure adopts 1-Max pooling, that is, a maximum feature is selected from the feature vectors generated by each sliding window, and then these features are spliced together to form a vector representation. You can also choose K-MAX pooling (select the largest K features in each feature vector), or average pooling (average each dimension in the feature vector), etc., the effect is to obtain a fixed-length vector representation of sentences of different lengths through pooling.

In summary, TextCNN (Convolutional Neural Network for Text Classification) has significant advantages in natural language processing, which are mainly reflected in the following aspects: First, TextCNN can effectively capture local features and patterns in text and automatically extract key information through convolutional operations, which makes it perform well in text classification tasks.

## 3. Methodology

Sentiment analysis has become particularly important in the age of social media and the Internet, allowing us to gain insight into the emotions, opinions and attitudes of the public in order to better understand social trends and public opinion. Tweets on social media platforms, as one of the main ways for users to express their emotions and opinions, contain a lot of valuable information. As a result, sentiment analysis of tweets has become an area of intense research, helping governments, businesses and research institutions to better understand the public's emotional leanings on specific issues.

### 3.1 Experimental introduction

In this context, this study aims to use TextCNN (Convolutional Neural Networks for Text Classification) for sentiment analysis of tweets. Our research question is: How can the TextCNN model be used effectively to analyse sentiment in tweets and how well does it perform? To answer this question, we will use a massive tweet dataset containing tweets from different social media platforms, covering a variety of topics and emotional categories.

TextCNN is a deep learning model known for its excellent performance in text classification tasks. It can capture local features and patterns in text, has a shallow network structure, is easy to train and deploy, and is suitable for processing text data of unlimited length. In this study, TextCNN is chosen as the main model to efficiently analyse the emotional tendency in large-scale tweet data.

In the following sections, we will detail key steps such as experimental design, data pre-processing, model training, and evaluation indicators to demonstrate our research methods and results. Our experimental hypothesis is that TextCNN can accurately analyse tweet sentiment, thus providing valuable insights for further research and application in the field of sentiment analysis.

### 3.2 Data preprocessing

Labeled sentiment analysis tasks can be treated as text subtasks. The dataset came from a course assignment SI650-Sentiment Classification at the University of Michigan and a dataset of tweets collected by Niek Sanders, with a total of about 157W tweets with a target variable of 0 or 1 indicating negative and positive emotions, and was a binary classification task for a large dataset. The paper mentioned that the naive Bayes classifier could achieve 75% accuracy. To verify this, I tested all the datasets using NB and SVM models respectively, and the accuracy of NB was 77.5% and that of SVM was 73%.

Code: from sklearn.feature_extraction.text import CountVectorizer

from sklearn.feature_selection import SelectKBest, chi2

from sklearn.linear_model import LogisticRegressionCV

from sklearn.pipeline import Pipeline

from sklearn.model_selection import train_test_split

### 3.4 Text model building

Steps:
① Use two layers of convolution
② Use more convolution nuclei, more scale convolution nuclei
③ BatchNorm is used
④ Two layers of full connection are used in the classification

Table : Model data set

| Concat | | | | | | | | |
|---|---|---|---|---|---|---|---|---|
| MaxPool | | MaxPool | | MaxPool | | MaxPool | | MaxPool |
| ReLU | | ReLU | | ReLU | | ReLU | | ReLU |
| BatchNorm | | BatchNorm | | BatchNorm | | BatchNorm | | BatchNorm |
| conv(1) | | conv(2) | | conv(3) | | conv(4) | | conv(5) |
| ReLU | | ReLU | | ReLU | | ReLU | | ReLU |
| BatchNorm | | BatchNorm | | BatchNorm | | BatchNorm | | BatchNorm |
| conv(1) | | conv(2) | | conv(3) | | conv(4) | | conv(5) |

Code: model.compile(loss='binary_crossentropy',

optimizer='adam',

metrics=['accuracy'])

history = model.fit(x_train_padded_seqs, y_train,

batch_size=32,

epochs=5,

validation_data=(x_test_padded_seqs, y_test))

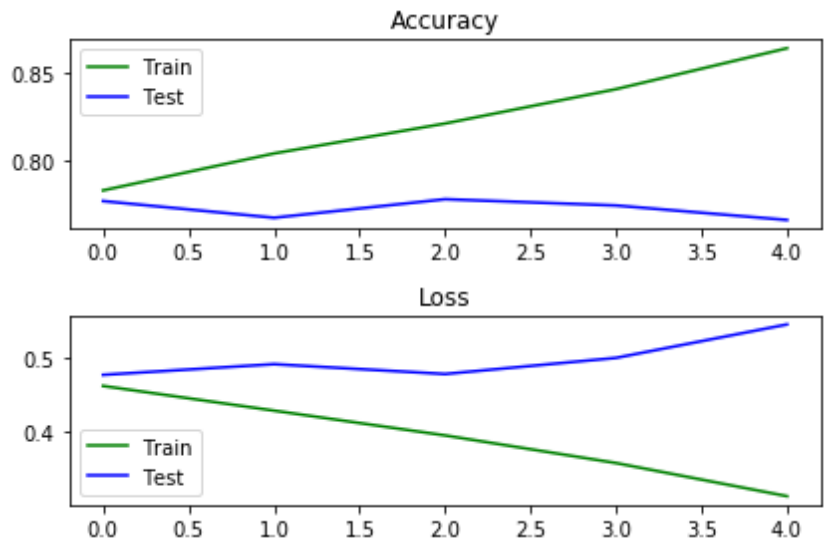

Figure 3: TextCNN result ROC curve

### 3.4 LSTM text model

RNN models can handle sequence problems, and LSTM is better at capturing long sequence relationships. Because of the existence of gate, LSTM can learn and grasp the dependencies in sequence well, so it is more suitable for dealing with long sequence NLP problems. The model structure is as follows:

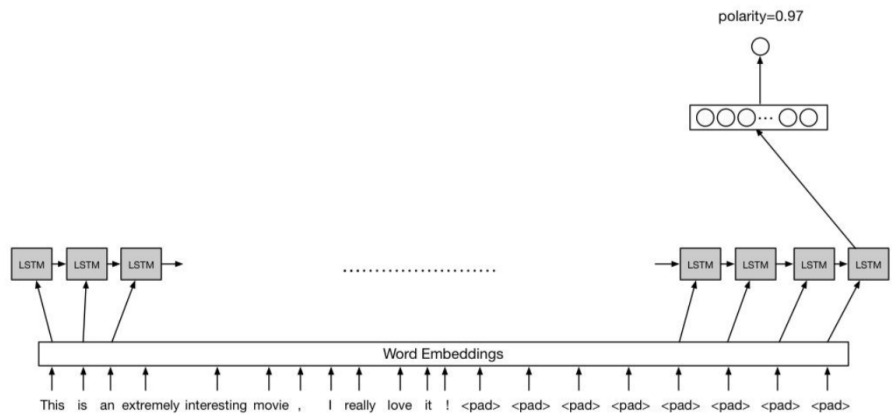

Figure 4: LATM text classification architecture diagram

Among them, the sentence is first word embedding, passed into the LSTM sequence for training, the last hidden state of LSTM is taken out, and the final output is obtained by adding the full connection layer.

When analyzing text classification under the LSTM model, it should be noted that in embeddings, unlike word vector summation in DNN, LSTM does not need to sum word vectors, but directly learns word vectors themselves. Whether it is summing or averaging, the convergent operation will lose some information

In the model, we first construct the LSTM unit and add dropout to prevent overfitting; After executing dynamic_rnn, we will get the final state of lstm, which is a tuple structure containing the cell state and hidden state (the result of passing through the output gate), we will only take the hidden state output here. lstm_state.h, concatenates this vector, and finally produces the output result.

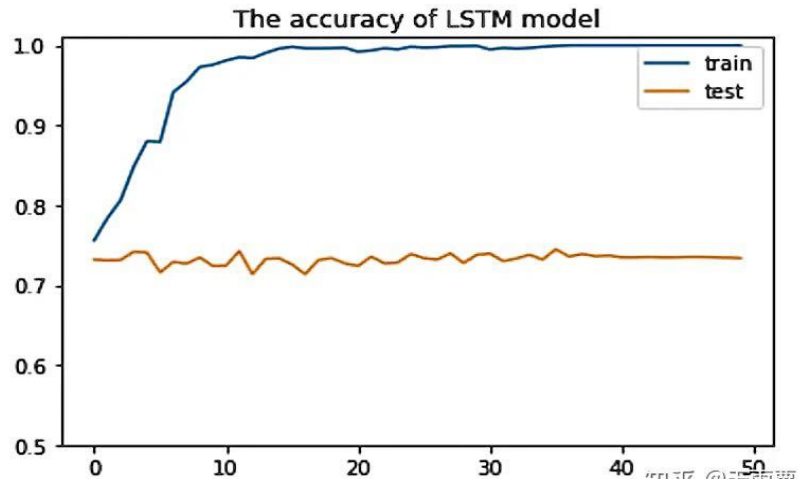

Figure 5： ROC curve of LATM text classification results

**3.5 Experimental result**

So far, we have completed the task of handling sentence classification using RNNS and CNNS respectively. Among them, the accuracy of DNN and RNN on test is almost the same, while the accuracy of CNN on test is 1%~2% higher, and the accuracy of multi-channels CNN on test is as high as 76.93%, and the training times are also less.

Our models are relatively simple, but generally speaking, these models have achieved good accuracy, which is largely due to the pre-trained word embedding, which shows the importance of word embedding in NLP models. The accuracy of multi-channels CNN is improved by adding task-specific information.

**Conclusion**

This paper discusses in detail the application of deep learning in natural language processing, especially in the field of text classification. Through the application of TextCNN and other models, it successfully realises the efficient processing of core tasks such as sentiment analysis. Deep learning has brought significant improvements to NLP technology, making the task of processing text data more flexible and efficient. The experimental results show that TextCNN and other models perform well in text classification tasks, especially under the pre-trained word embeddedness, and achieve satisfactory performance. In addition, this paper also focuses on the technical challenges of text generation, text classification and semantic parsing, and proposes an integrated interactive training method to effectively address these challenges, which provides an important reference for further research and application in the field of NLP.

As one of the most important tasks in natural language processing, text classification is constantly evolving and expanding. In the future, with the continuous development and improvement of deep learning technology, text classification will bring more opportunities and challenges. Some of these future directions are

1. Multimodal text classification: Combining text with multimodal information such as images and audio for classification, so that text classification can better meet the processing needs of multimedia content.

2. Cross-lingual text classification: Develop a text classification model that can handle multiple languages, and promote the analysis of multilingual social media and international information.

3. Continuous learning: Develop text classification models that can continuously learn and adapt to new data and domains to cope with the changing information environment.

4. Privacy protection: Strengthen the protection of user privacy in text classification to prevent the misuse of personal information.

In summary, text classification has a wide range of application prospects in natural language processing, which is not only of great value in public opinion analysis, information retrieval and other fields, but also plays an important role in everyday life, such as social media and intelligent assistants. As technology continues to advance, we can expect text classification to play a greater role in improving natural language processing techniques and solving practical problems.